\documentclass[letterpaper, 10 pt, conference]{ieeeconf}  %

\IEEEoverridecommandlockouts                              %

\usepackage{graphics} %
\usepackage{epsfig} %
\usepackage{mathptmx} %
\usepackage{times} %
\usepackage{amsmath} %
\usepackage{amssymb}  %

\usepackage{ifthen}
\usepackage{subcaption}
\usepackage[dvipsnames]{xcolor}
\usepackage{hyperref}
\usepackage{gensymb}
\usepackage{lineno}
\usepackage[T1]{fontenc}
\usepackage{textcomp}
\usepackage{fancyhdr}
\usepackage{graphics}
\usepackage{epsfig}
\usepackage{mathptmx}
\usepackage{times}
\usepackage{soul}
\usepackage{cite}
\usepackage{url}

\usepackage{breakurl}
\definecolor{red}{rgb}{255,0,0}

\title{\LARGE \bf Set Phasers to Stun: Beaming Power and Control\\to Mobile Robots with Laser Light}

\author{Charles J. Carver$^{\dag2}$, Hadleigh Schwartz$^{\dag1}$, Toma Itagaki$^{1}$, Zachary Englhardt$^{3}$, Kechen Liu$^{1}$, \\ Megan Graciela Nauli Manik$^{1}$,  Chun-Cheng Chang$^{2}$, Vikram Iyer$^{2}$, Brian Plancher$^{4}$, and Xia Zhou$^{1}$ %
\thanks{$^\dagger$Co-primary authors }
\thanks{$^{1}$ Department of Computer Science, Columbia University}%
\thanks{$^{2}$ Lincoln Laboratory, Massachusetts Institute of Technology (prev. 1)}%
\thanks{$^{3}$ Paul G. Allen School of Computer Science \& Engineering, University of Washington}
\thanks{$^{4}$ Barnard College, Columbia University}
\thanks{Corresponding author: {\tt\footnotesize \{hadleigh\}@cs.columbia.edu}}%
}

\fancypagestyle{firstpage}{
    \fancyhf{} 
    \fancyfoot[C]{\small~\copyright2025 IEEE.  Personal use of this material is permitted.  Permission from IEEE must be obtained for all other uses, in any current or future media, including reprinting/republishing this material for advertising or promotional purposes, creating new collective works, for resale or redistribution to servers or lists, or reuse of any copyrighted component of this work in other works.} 
}

\begin{document}

\newcommand{\name}{Phaser}
\newcommand{\figr}[1]{{Figure~\ref{fig:#1}}}
\newcommand{\tabr}[1]{{Table~\ref{tab:#1}}}
\newcommand{\para}[1]{{\noindent{\bf \noindent #1 }}}
\newcommand{\citeme}[1]{{\color{red}[CITE]}}

\maketitle
\thispagestyle{firstpage}
\pagestyle{empty}

\begin{abstract}

We present \name, a flexible system that directs narrow-beam laser light to moving robots for concurrent wireless power delivery and communication. We design a semi-automatic calibration procedure to enable fusion of stereo-vision-based 3D robot tracking with high-power beam steering, and a low-power optical communication scheme that reuses the laser light as a data channel. We fabricate a \name\ prototype using off-the-shelf hardware and evaluate its performance with battery-free autonomous robots. \name\ delivers optical power densities of over 110 mW/cm$^2$ and error-free data to mobile robots at multi-meter ranges, with on-board decoding drawing 0.3~mA (97\% less current than Bluetooth Low Energy). We demonstrate \name\ fully powering gram-scale battery-free robots to nearly 2x higher speeds than prior work while simultaneously controlling them to navigate around obstacles and along paths. Code, an open-source design guide, and a demonstration video of \name\ is available at \texttt{\textbf{\url{https://mobilex.cs.columbia.edu/phaser/}}}. 
\end{abstract}

\section{Introduction}

Mobile, autonomous robots play an increasingly important role in today's world, with the potential to perform tasks in warehouses, factories, and homes and conduct advanced environmental explorations ~\cite{kabir2023internet}. However, the significant \emph{power} needed for locomotion, on-board computation, and communication presents a key barrier to the broader deployment of such robots. Given the energy density of current batteries~\cite{thackeray2012electrical}, most autonomous robots today either remain tethered by charging wires or must routinely return to charging stations, reducing deployment time. This problem is exacerbated in miniaturized robots, which cannot support the 100s of milligrams of battery payload~\cite{ma2015design, james2018liftoff, fuller2019four, bhushan2018milligram, jafferis2019untethered} needed for extended operation, even on their milliwatt power budgets.

\begin{figure}[t]
    \centering
    \includegraphics[width=0.9\columnwidth]{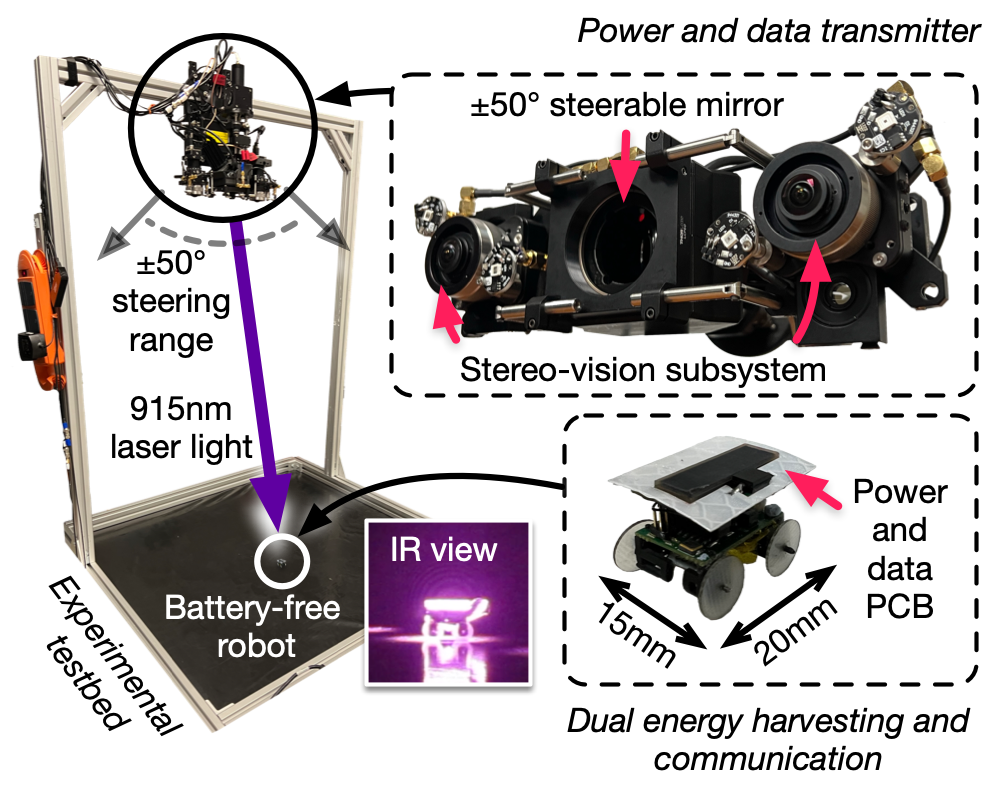}
    \caption{\name\ utilizes narrow-beam laser light for simultaneous power delivery and communication with mobile robots. It consists of a power and data transmitter that utilizes stereo-vision to perform 3D localization and steer light to the robot as it is on the move. The robot is equipped with a retroreflective marker to ease tracking, a solar cell for power harvesting, and a photodiode to decode control commands from the laser light.}
    \label{fig:teaser}
\end{figure}

To address this challenge, recent robotic systems have turned to \emph{wireless} power delivery, where radio frequency~\cite{johnson2023millimobile, ozaki2021wireless} or, more commonly, optical energy~\cite{elkunchwar2021toward, jafferis2019untethered, johnson2023millimobile, shen2024sunlight, liller2025development, james2018liftoff, bhushan2018milligram} is provided remotely and harvested by on-board robots. The majority of such systems utilize diffuse light, resulting in major inefficiencies and unrealistic deployment requirements as optical power is diluted over a wide area (e.g., 3 suns of optical output to power a robot less than a meter away~\cite{jafferis2019untethered}, geographical and seasonal operation constraints~\cite{liller2025development}). A more efficient, environment-agnostic approach is to utilize directional laser light to deliver concentrated optical power. However, this introduces a technical challenge in continuously tracking the robot and steering the narrow beam to its optical receivers. This task is non-trivial as it requires precisely integrating 3D target tracking with laser steering, in the face of the stringent steering range, target speed, and optical energy density requirements of mobile power delivery. Most prior demonstrations have not tackled this challenge and were thus limited to a few seconds of operation~\cite{james2018liftoff, bhushan2018milligram}, after which the robot moved out of the static laser beam.

As such, we present \name, a flexible system framework that directs narrow-beam laser light to moving robots for concurrent power delivery and data communication. At the core of \name\ is a vision-based tracking and steering system that maintains alignment of high-power laser beams with mobile targets for power and data transmission. Unlike conventional approaches constrained by diffuse light and environmental conditions, \name\ supports a range of laser wavelengths, optical powers, steering devices, tracking algorithms, robot characteristics (e.g., size, number, motion characteristics), and deployment configurations. %

Our key contributions are as follows. (1) We design a low-overhead calibration mechanism for integrating 3D tracking via stereo-vision with any laser beam steering mechanism and realize its integration with a high-power, large-field-of-view steering system. We deliver optical power densities  of over 110~mW/cm$^2$ (greater than one sun\footnote{One sun is approximately 100~mW/cm$^2$~\cite{sun_irradiance}}) with a standard deviation of only 1.9 mW/cm$^2$ across robot locations in three dimensions. Our optical circuit minimizes energy losses to $\leq$9\%, achieving a 133\% increase in optical power throughput compared to prior work~\cite{carver2024catch}. (2) We design a low-power communication architecture and receiver circuit that draws only 0.3~mA of current (less than 6.7\% of that required for battery-free robot locomotion~\cite{johnson2023millimobile}) to decode frequency-shift keyed signals. It achieves a $\leq10^{-3}$ bit error rate for laser incident angles up to $\pm90\degree$ and depths up to 5~m, ensuring reliable data reception across diverse conditions. (3) We integrate \name\ with MilliMobiles~\cite{johnson2023millimobile} -- gram-scale battery-free robots -- and demonstrate robot operation powered and controlled via laser light to locomote around obstacles and along paths.  %
\section{Related Work}
\label{sec:related}

\para{Power Delivery to Robots}
To avoid the wired \emph{tethers} required by many robotic platforms~\cite{chen2019controlled, wu2019insect, hutama2021millimeter}, recent works have increasingly examined \emph{wireless} power delivery, wherein radio frequency (RF) ~\cite{johnson2023millimobile, ozaki2021wireless} or optical energy~\cite{elkunchwar2021toward, jafferis2019untethered, johnson2023millimobile, shen2024sunlight, liller2025development, james2018liftoff, bhushan2018milligram} is provided remotely and harvested on robots. Optical approaches are particularly promising because light can be focused to near points, achieving uniquely high energy densities. In contrast, RF beams are physically constrained by antennae and dish size, limiting achievable energy density.

Optical power delivery works can be separated into undirected and directed designs. In the former, diffuse light (e.g., from LEDs or the sun) is used to deliver power over a wide area~\cite{elkunchwar2021toward, jafferis2019untethered, johnson2023millimobile, shen2024sunlight, liller2025development}, resulting in unrealistically high light level requirements (e.g., 3 suns~\cite{jafferis2019untethered}) and stringent limitations on payload, speed, or environmental conditions~\cite{elkunchwar2021toward, jafferis2019untethered, johnson2023millimobile, shen2024sunlight, liller2025development}. In directed optical power systems, narrow-beam laser light is employed~\cite{blackwell2005recent, james2018liftoff, jin2018wireless, iyer2018charging, sheng2022adaptive, bhushan2018milligram, 12hflight2011lasermotive}. This boosts delivered energy density but requires continuous robot tracking and laser steering to maintain beam alignment with onboard solar cells. Because of the difficulty of this task (elaborated in Section \ref{subsec:track_steer_design}), most directed-light works support only stationary targets~\cite{iyer2018charging, sheng2022adaptive}, require manual beam steering~\cite{blackwell2003nasa}, or demonstrate only seconds of mobility~\cite{james2018liftoff, bhushan2018milligram}. Outside the domain of battery-free and autonomous robots, laser-based power delivery has remained an experimental approach largely focusing on space systems and flight time extension, with only a few works showing promising results~\cite{blackwell2003nasa, 12hflight2011lasermotive}.

\para{Target Tracking and Laser Steering} Many prior works study integrating 3D object tracking and laser steering for mobile targets. Most existing systems require equipping the target with active electronics like LEDs~\cite{milanovic2011memseye}, laser beacons~\cite{sofka2009laser,  llcdpat}; RF transceivers~\cite{abadiautolocatingfso, ishola2021pat, gupta2022cyclops}; or photodiodes~\cite{ishola2021pat}, which offset the gains of delivered optical power. Other works \cite{kasturi2016uav, milanovic2011memseye} support equipping the target with retroreflectors --- passive elements -- but use iterative scans for beam alignment, causing periodic breaks in connection. 

In contrast, recent work~\cite{carver2024catch} develops an optical design that colocates a steerable laser with a single image sensor used for generic vision-based  object tracking, achieving continuous beam alignment with high-velocity targets. However, \cite{carver2024catch} can only support low power densities of 10s of mW/cm$^2$ as a result of the low damage thresholds of its core optical components. ~\cite{carver2024demonstration} explores a simplified version of this design for power delivery but is limited to targets moving in a 2D plane.
Existing systems supporting 3D motion \emph{and} high optical powers (i.e. Watts)~\cite{roberts2016overview, abdelfatah2022review, llcdpat} utilize bulky gimbals tailored for kilometer-range targets with low angular speeds and fixed trajectories (e.g., satellites moving at $1.5\degree$/s~\cite{walsh2022demonstration}). This is inadequate for applications involving robots moving with arbitrary trajectories at meter-level ranges.

\section{\name\ Design and Implementation}
\name's design consists of two core elements: a) a stereo-vision-based robot tracking and laser steering system, and b) a low-power optical communication scheme and receiver to reuse laser light for data transmission. These elements enable \name\ to continuously deliver power and data to robots moving freely in 3D space at meter-level ranges.

\subsection{Stereo-Vision Tracking and High-Power Beam Steering}
\label{subsec:track_steer_design}
Maintaining beam alignment with mobile robots is key to effective laser-based power delivery and communication. Even a 1~cm error in the end location of a steered laser beam can cause light to completely miss a robot's optical receivers.

\begin{figure*}[t]
    \centering
     \begin{subfigure}[t]{0.26\textwidth}
        \centering
        \includegraphics[width=1\textwidth]{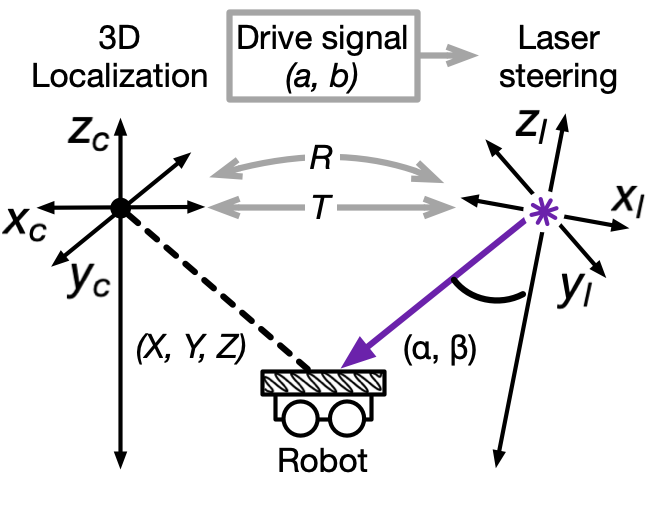}
        \caption{Geometry and unknowns.}
        \label{fig:track_steer_chall}
    \end{subfigure}\hfill
    \begin{subfigure}[t]{0.28\textwidth}
        \centering
        \includegraphics[width=0.98\textwidth]{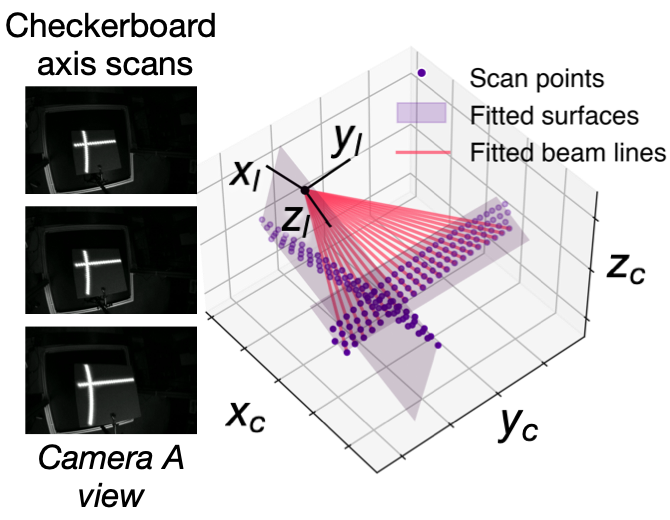}
        \caption{Calibration of $R$ and $T$.}
        \label{fig:pose_recovery}
    \end{subfigure}\hfill
    \begin{subfigure}[t]{0.3\textwidth}
        \centering
        \includegraphics[width=0.98\textwidth]{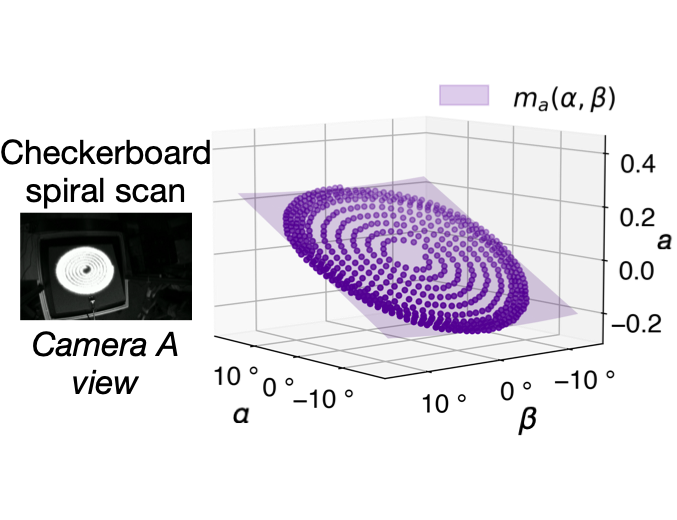}
        \caption{Calibration of steering drive signal $a$.}
        \label{fig:mapping}
    \end{subfigure}
    \caption{Stereo-vision tracking and steering design. (a) To convert 3D robot locations to outgoing laser angles ($\alpha, \beta$), the system must know the steering and localization devices' relative rotation $R$ and translation $T$, as well as the precise steering device drive signals required to produce ($\alpha, \beta$). \name's calibration recovers these quantities from a set of short, automated scan sequences on the calibration checkerboard (visualized in timelapse images pulled from a stereo-camera). (b) Triangulated scan points are used to recover the steering device origin and axes in the stereo-pair's reference frame,  equivalent to $T$ and $R$. (c) A spiral of scan points is fitted to mapping functions $m$, which output the drive signals $(a, b)$ needed to steer at a desired angle $(\alpha, \beta)$.}
    \label{fig:calibration}
    \vspace{-10pt}
\end{figure*}

This task raises two challenges. First, as illustrated in Figure \ref{fig:track_steer_chall}, integrating 3D tracking with laser steering requires precise calibration of the localization and steering devices' relative translation and rotation, so that a 3D robot position can be converted to an outgoing 2D laser angle. However, obtaining these measurements is non-trivial, as the origins of these devices typically lie at unknown points, and the orientations of their axes are often unspecified. Complicating this matter, beam steering systems take as input drive signals (e.g., mirror actuation voltages or unit-less commands) with often non-linear relationships to the resulting outgoing angle of laser light. Second, typical tracking and laser steering optical elements (e.g., wide-angle lenses for expanding steering field-of-view\cite{milanovic2011memseye}, beamsplitters for alignment~\cite{milanovic2011memseye, ishola2021pat}) introduce power losses and beam divergence that can dramatically reduce delivered energy. For example, the five-element design in~\cite{carver2024catch} incurs a 61\% loss in optical energy density. While some systems mitigate these losses by using gimbals to rotate laser and tracking hardware as a rigid unit, they possess limited steering speeds unsuitable for maintaining alignment with fast-moving robots (Section~\ref{sec:related}).

To solve these challenges, we develop a flexible framework for integrating 3D object tracking via stereo-vision with real-time beam steering, underpinned by a low-overhead calibration procedure. The framework is agnostic to steering architecture, laser configuration, and visual tracking algorithm, affording maximal flexibility in our system design. We additionally design a steering mechanism optimized for large-field-of-view (FoV) steering of multi-Watt beams, accommodating our application's mobility and power demands.

The tracking and laser steering system (Figure \ref{fig:prototype}) co-locates the stereo-cameras (Alvium 1800 U-240m) and beam steering hardware. It incorporates a 915 nm laser diode with variable power up to 12~W (RPMC 915nm 12W Fiber Coupled Diode Laser). This wavelength is chosen as commercial solar cells have optimal conversion efficiencies in the near-infrared region. The laser is powered by a Rohde \& Schwarz HMP4000 power supply, and the entire system is controlled via an Apple M1 Max with 64~GB of RAM.

\para{Low-Overhead Calibration} Our calibration procedure recovers each of the unknowns necessary for integrating 3D tracking with 2D laser steering (Figure~\ref{fig:track_steer_chall}), namely the $R$ and $T$ matrices and the relationship between steering drive signals and outgoing laser angles. It augments traditional checkerboard-based stereo-camera calibration techniques~\cite{tsai1987versatile} with two automated laser scan sequences. The entire procedure takes as input a single video of the checkerboard as it is moved and scanned, as captured by the stereo-camera, and can be broken down into three logical stages.

First, we capture frames of the checkerboard at several positions and angles. With these frames, \name\ solves for the stereo-camera parameters using OpenCV~\cite{OpenCV}, after which it is able to perform 3D localization using triangulation~\cite{hartley1997triangulation}. %

Second, we solve for $R$ and $T$ using a judiciously collected set of 3D points from our newly calibrated stereo-cameras and several geometric guarantees afforded by our steering device. We begin by recovering $R$, which is simply a matrix defined by the steering axis directions $\vec{x_l}, \vec{y_l}, \vec{z_l}$. As shown in Figure~\ref{fig:pose_recovery}, we fix the checkerboard at three positions\footnote{A minimum of two positions are needed; additional positions increases accuracy by providing a larger sample of points.}. At each, we scan along the steering device's x- and y-axes by using steering inputs of the form $(a_i, 0)$ and $(0, b_i)$. We detect the beam's incident points on the board via contour detection (described further below) and  triangulate them to obtain a set of 3D points for each axis. Then, we fit degree-2 polynomial surfaces $SX_Z$ and $SY_Z$ to each set to accommodate for any distortion-induced curvatures in the points~\cite{distortion_optotune_mre3_manual}. Intuitively, the surfaces' intersection defines $z_l$. Thus the steering device's XZ and YZ planes are simply the planes tangent to $S_{XZ}$ and  $S_{YZ}$, respectively, at $z_l$. We can thus compute the XZ-plane as: 
\[
    z = S_{XZ}(x_0, y_0) + \left. \frac{\partial S_{XZ}}{\partial x} \right|_{(x_0, y_0)} (x - x_0) + \left. \frac{\partial S_{XZ}}{\partial y} \right|_{(x_0, y_0)} (y - y_0),
\]
where $(x_0, y_0, z_0)$ is any point on  $z_l$. The YZ-plane is computed via the same equation using $S_{YZ}$ instead of $S_{XZ}$.

Then, $\vec{x_l}$, satisfies the following system of equations
\begin{equation*}
\begin{aligned}
    \vec{x_l} \cdot \vec{n}_{XZ} &= 0 \quad \quad \vec{x_l} \cdot \vec{z_l} &= 0,
\end{aligned}
\end{equation*}
where $\vec{n}_{XZ}$ is the normal vector to the XZ plane. The vector $\vec{y_l}$ is computed similarly using the YZ-plane normal.

Recovering $T$ requires determining the exact position of the origin along $z_l$. Here, we know that when the laser beam is steered at a fixed outgoing angle, any points that it passes through define a line with its start at the steering device’s origin. Thus, the scan points can be grouped by outgoing angles and fitted to lines representing these "virtual beams" (visualized in pink in Figure \ref{fig:pose_recovery}). The intersection of all these lines, given by a system of linear equations, is the origin. 

In the third and final stage, we establish mapping functions relating drive signals to resulting outgoing angles of laser light (Figure~\ref{fig:mapping}). To do so, we fix the checkerboard and instruct the steering device to scan in a spiral pattern. For each input of the form $(a, b)$, we record the 3D position of the laser spot upon the board and convert it to a laser angle $\alpha$ and $\beta$ using $R$, $T$, and our calibrated stereo-vision system. We then fit two functions --- one to our set of $(\alpha, \beta, a)$ data points and another to the $(\alpha, \beta, b)$ points -- using a multivariate, polynomial regression. This yields:
\[
    m_a(\alpha, \beta) = \sum_{i=0}^{2}  \sum_{j=0}^{2} m_{ij} \alpha^i \beta^j, 
    \quad
    m_b(\alpha, \beta) = \sum_{i=0}^{2}  \sum_{j=0}^{2} n_{ij} \alpha^i \beta^j,  
\]
where the coefficients $m_{ij}$ and $n_{ij}$ are determined by the respective regressions. Since $a$ and $b$ each may affect outgoing angles along both dimensions, both $\alpha$ and $\beta$ are utilized in the regressions for $m_a$ and $m_b$. We choose a second-degree polynomial to account for mirror actuation non-linearities.

\para{Real-Time 3D Tracking and Steering} During deployment, \name\ tracks and steers to mobile robots in real-time. For each pair of images produced by its stereo-cameras, it detects the robot using any vision-based detection technique, obtains a 3D position via triangulation, and transforms the position via $R$ and $T$ to obtain $(\alpha, \beta)$. \name\ then queries the mapping functions to determine the necessary steering device inputs. %

Our implementation leverages retroreflectors to track the robot in the stereo-camera stream. This affords high detection speeds using simple computer vision algorithms. Specifically, we equip targets with a 0.5~g retroreflective tag with a cutout to house the solar cell (shown on our robot in Figure~\ref{fig:prototype}).
When \name\ illuminates the scene with wide-angle LEDs co-located with each stereo-camera (Thorlabs M810D4), the retroreflective tag registers in images as a bright blob with a center corresponding to that of the solar cell. We perform Otsu binarization and contour detection~\cite{hoff1989surfaces} on images to determine these center pixel locations, which serve as the triangulation input. To increase the SNR of robot blobs and ensure that the laser's beam spot does not register as a confounding blob, we slide optical filters (Everix Ultra-Thin OD4) in front of the image sensors during deployment. 

\para{High-Power, Large FoV Beam Steering} Our beam steering design consists of the minimal number of optical elements for fast steering in a $\pm 50\degree$ range, simultaneously accommodating high optical power throughput and robot mobility.

At the core of the optical design is a voice coil-actuated mirror with a $\pm 25\degree$ mechanical tilt range and ${\sim}$300~Hz steering rate (Optotune MR-15-30). This steering architecture enables fast point-to-point steering with large angular ranges and high precision~\cite{kim2024design}. In our optical circuit, laser light first passes through an aspheric lens which collimates the beam to produce a spot 2.4~cm in diameter at a distance of 1.3~m, thus tightly circumscribing the robot's receiver photovoltaics in our testbed (Section \ref{subsec:exp_setup}).
Light is then directed onto the steerable mirror by a fixed mirror (Thorlabs 1/2" x 1/2" Protected Gold Mirror) oriented at $39\degree$ from the steerable mirror's z-axis. This eliminates clipping or blind spots that would result if the laser diode were positioned to exit directly onto the steerable mirror. The optical components are housed in a 3D-printed enclosure maintaining the desired positions. 

With only a focusing lens and two mirrors between the laser diode and steerable mirror's exit point, we have maximal control over beam divergence and achieve power throughput constrained primarily by the mirrors' reflectance ratios. At the 915~nm wavelength of our prototype's laser, the focusing lens has a transmission ratio of 99.8\%, and the gold-coated  and steerable mirrors have reflectance ratios of 96\% and 95\%, respectively, for an overall loss of only 8.9\%.

\begin{figure}[t]
    \centering
    \includegraphics[width=0.8\columnwidth]{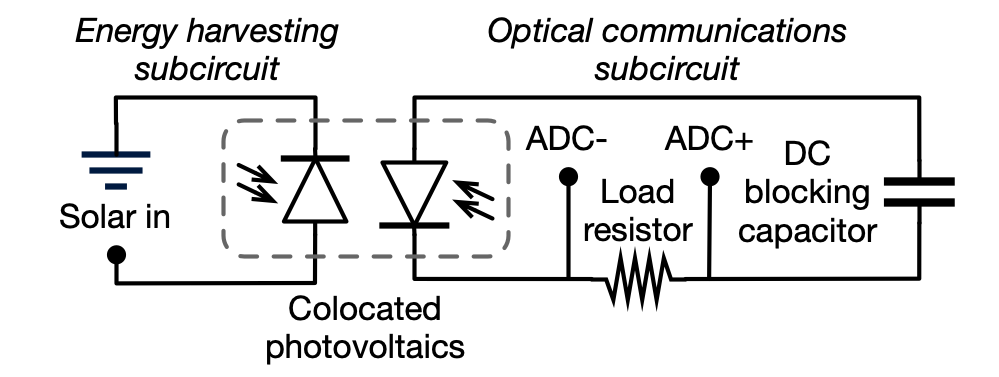}
    \caption{Analog circuit used to simultaneously harvest optical power and receive optical data via a co-located solar cell and photodiode.}
    \label{fig:circuit}
    \vspace{-15pt}
\end{figure}

\subsection{Low-Power Optical Communication}
Our low-power communication scheme reuses the optical power medium as a communication channel. We modulate the high-power laser light with low-current signals that can be decoded with simple hardware and algorithms consuming only 0.3~mA of current -- significantly less than that harvested from the laser itself. This is just ${\sim}$3\% of the current required for Bluetooth Low Energy reception (Section~\ref{subsec:comms_eval}).

To efficiently modulate the high-power laser beam, we choose frequency-shift keying (FSK), which transmits bits by modulating the carrier signal with dedicated frequencies~\cite{jeromin1986m}. FSK is robust to amplitude variation and does not require precise pulse timing unlike alternative optical schemes, such as pulse position modulation. We choose 12.5~Hz and 25~Hz as our FSK frequencies and modulate the laser directly via the power supply, which is controlled via its serial interface. 

\begin{figure}[t]
    \centering
    \includegraphics[width=0.9\columnwidth]{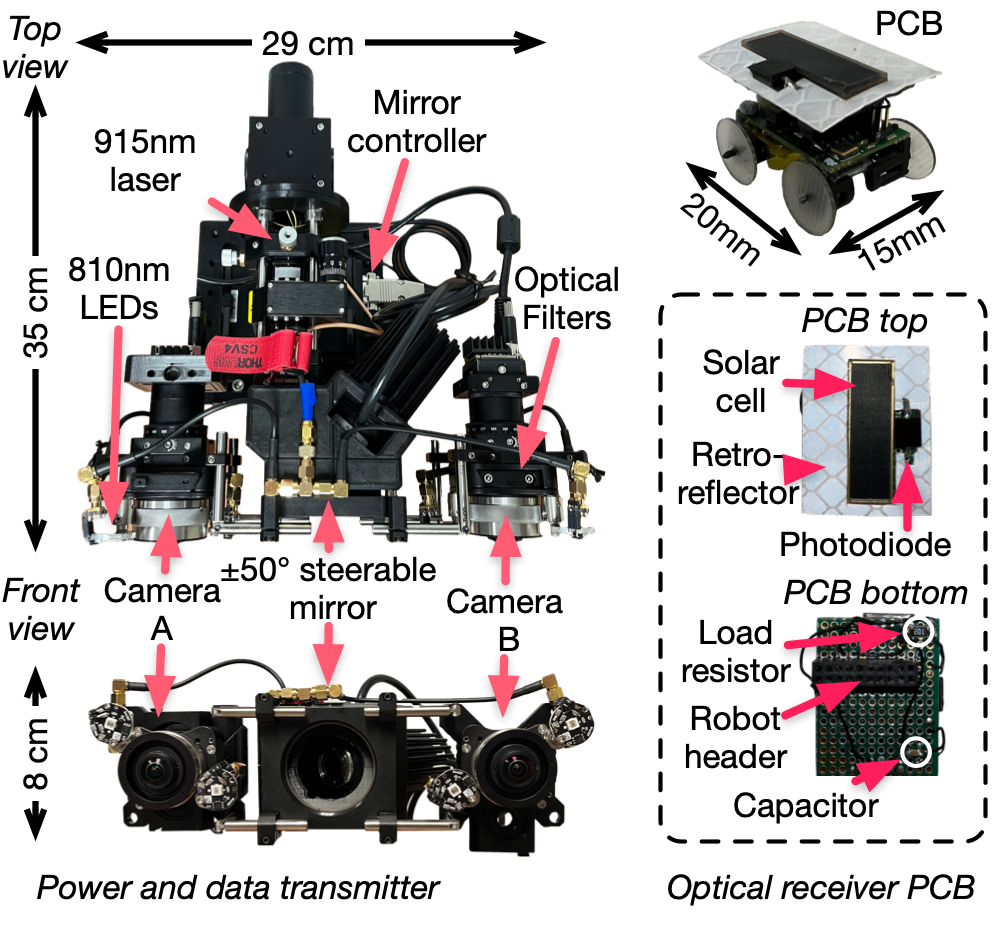}
    \caption{ The \name\ prototype. The tracking and steering system (left) consists of a stereo-pair and a $\pm50\degree$ steerable mirror. Four 810~nm light-emitting diodes support retroreflective imaging of the robot, and a 915~nm fiber-coupled laser with optical power up to 10~W is used for power delivery and communication. Onboard robots, a custom PCB enables energy harvesting and data reception and a retroreflective tag serves as a tracking marker.}
    \label{fig:prototype}
    \vspace{-15pt}
\end{figure}

Because solar cells used for energy harvesting are often connected to supercapacitors~\cite{johnson2023millimobile, bhushan2018milligram}, which act as low-pass filters, it is infeasible to reuse the solar cell as our communication receiver. Thus, we co-locate a photodiode (Vishay Semiconductors - BP104)  with the solar cell 	(ANYSOLAR KXOB25-01X8F-TR) (Figure~\ref{fig:prototype}). 
As our collimated laser produces circular beam spots, utilizing an additional photodiode incurs no overhead; a beam steered to the center of the retroreflector will fully circumscribe both receivers.

To receive the encoded signals via the photodiode, we design a simple resistor-capacitor circuit (Figure~\ref{fig:circuit}). The photodiode is unbiased and thus operates in photovoltaic mode, reducing the power consumption of the communication channel. It is connected to a 470~k$\Omega$ resistor and 1~uF capacitor. The latter removes the direct current component of the received laser signal, passing only the FSK-modulated alternating current component. The signal is
read at a robot's microcontroller via an analog-to-digital converter (ADC). 

The demodulation algorithm running on the microcontroller utilizes the timing between signal peaks to recover transmitted FSK frequencies, which directly correspond to a symbol (e.g., 0 or 1 bit). This approach produces accurate results while removing the necessity for a Fast Fourier Transform for frequency computation, which is both computationally expensive and plagued by component frequencies when decomposing square wave FSK signals.

\section{Evaluation}
\label{subsec:impl}

\subsection{Experimental Setup and Metrics}
\label{subsec:exp_setup}

We utilize a 0.91~m $\times$ 0.91~m testbed with the prototype placed facing downwards at an elevation of 1.3~m (Figure \ref{fig:teaser}) and calibrate as described above. The ground material is a medium-hard neoprene rubber. Ambient light levels were held constant at ${\sim}$600~lx. We operate the laser with 6.3 W of electrical power, which consistently generated a minimum 110 ~mW/cm$^2$ irradiance for targets at 1.3~m.

We utilize second-generation MilliMobiles, differing from initial versions~\cite{johnson2023millimobile} in their upgraded microprocessor (nRF5340-series Bluetooth SOC). We modify the robots only to add our retroreflector and circuitry (Figure \ref{fig:prototype}) and adjust the firmware to run decoding. The ADC is set to sample at 100~Hz, and we map 12.5~Hz, 25~Hz, and 0~Hz frequencies to left, right, and forward steering commands.  The robot discharges either its left or right motor to realize a turn.

When assessing received laser power, we measure the \emph{optical power density}, i.e., the irradiance (mW/cm$^2$). This metric quantifies the density of laser energy \name\ can deliver to a robot's onboard optical receivers. 

For assessing communication performance, we first compute the relationship between the communication signal's signal-to-noise ratio (SNR) and link bit error rate (BER). SNR is the ratio between the ADC's peak-to-peak range ($R$) when the photodiode is receiving light at a given intensity and the experimentally measured noise floor ($N$) of 8~mV. BER measures the percentage of incorrectly-decoded bits among the total received bits. Then, in each experiment, we measure the communication link SNR and report its corresponding BER based on the above relationship. Using SNR as the anchor of our communication experiments abstracts the impacts of wavelength, laser power, environmental conditions, or photodiode model; any optical source capable of generating a given SNR under its deployment configuration should expect the BER given by our SNR-BER relationship.

\para{Ensuring Laser Safety} To comply with laser safety requirements, experiments are conducted in a room with only investigators present, wearing appropriate safety goggles.

\subsection{Power Delivery Evaluation}
\label{subsec:power_eval}

\para{Overall Performance} 
We evaluate \name's performance in delivering power based on the received irradiance across target locations in 3D space. %
With perfect tracking and steering, the power received at all equidistant points radially surrounding the laser's origin \emph{and} at a fixed depth should be roughly equal. Further, as target depth increases, the irradiance may decrease due to attenuation or beam divergence but should remain equal among all points at the depth. To assess \name's conformance to these standards, we create a board with eight points spaced 24~cm apart in a grid centered with the laser's origin. We then place a Thorlabs PM400 optical power meter equipped with a retroreflective tag at each point and record an average irradiance at each across three trials for three grid depths $z$. Because this experiment requires taking manual measurements in the proximity of the laser beam, we adjust the laser electrical power to 3~W for safety. This results in lower irradiance values than in other experiments.

\begin{figure*}[t]
\vspace{4pt} %
\begin{subfigure}[t]{.3\textwidth}
    \centering
    \hfill
    \begin{subfigure}[t]{0.34\columnwidth}
        \centering
        \includegraphics[width=1\textwidth]{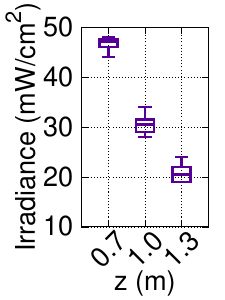}
    \end{subfigure}
    \begin{subfigure}[t]{0.63\columnwidth}
        \centering
        \includegraphics[width=1\textwidth]{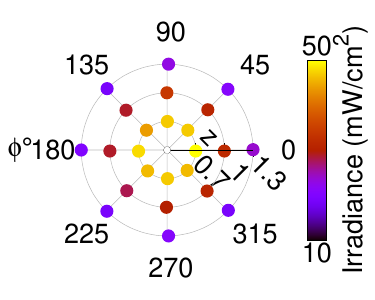}
    \end{subfigure}
    \caption{Irradiance w/ 3W laser electrical power.}
    \label{fig:accuracy}
\end{subfigure}
\hfill
\begin{subfigure}[t]{.3\textwidth}
    \centering
    \includegraphics[width=1\columnwidth]{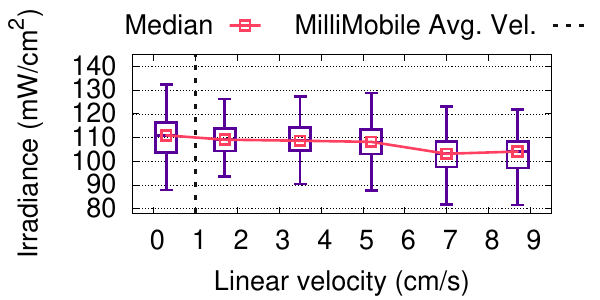}
    \caption{Effect of robot velocity.}
    \label{fig:speed}
\end{subfigure}
\hfill
\begin{subfigure}[t]{.3\textwidth}
    \centering
    \includegraphics[width=1\columnwidth]{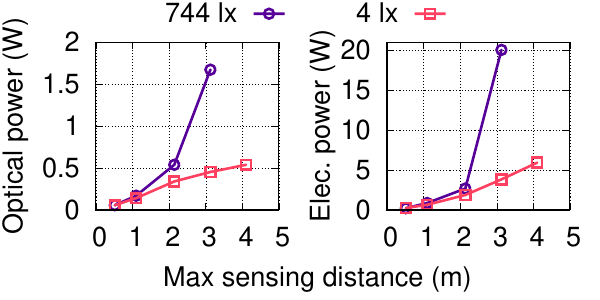}
    \caption{Required LED power by robot distance.}
    \label{fig:led-depth}
\end{subfigure}
\caption{Power delivery performance results, including (a) received irradiance under adjusted laser power for eight different target azimuths $\phi$ relative to the laser origin and three depths, (b) received laser irradiance as a function of robot velocity, and (c) required electrical and optical LED power to sense the robot at different distances via retroreflective imaging, under two ambient light conditions (744 lx and 4 lx).}
\vspace{-10pt}
\end{figure*}

Figure \ref{fig:accuracy} shows a boxplot of irradiances obtained at $z=$0.7, 1.0, and 1.3~m, and a polar plot illustrating the distribution of irradiances around the origin. Across all measurements, we observe a standard deviation of only 1.9~mW/cm$^2$, indicating the laser light is reliably delivered to each 3D location regardless of location. The per-depth standard deviations of irradiance are equally low, at 2.00~mW/cm$^2$, 1.93~mW/cm$^2$, and 1.91~mW/cm$^2$, respectively. 

\para{Robot Velocity} Supported robot velocity is primarily constrained by stereo-vision tracking latencies as well as the steerable mirror's response (${\sim}$300~Hz for small-angle steps). To understand the impact of these delays, we place a  20~mm-long rotating arm below the prototype, aligned with the laser origin. We attach the optical power meter (again enclosed by the retroreflective tag) to the end of arm, rotate it at fixed angular velocities using a programmable motor, and continuously sample the irradiance for 10 arm rotations. 

Figure \ref{fig:speed} shows a boxplot of recorded irradiance values per tested linear velocity, computed from angular velocity via the rotating arm's length. Given the confounding impacts of laser incidence angle, we highlight the median irradiance at each speed. At the slowest and fastest speeds of 0.3~cm/s and  8.7~cm/s, we observe median irradiances of 111~mW/cm$^2$ and 104~mW/cm$^2$, respectively, corresponding to a 6.3\% decrease. The grey dashed line denotes the average speed of the MilliMobiles in our integration tests (Section \ref{subsec:end_to_end}), with a 1.8\% decrease compared to the 0.3~cm/s case.

\para{Robot Distance} The maximum supported robot distance is determined by the stereo-vision tracking range and the laser's irradiance as a function of distance. Regarding the latter, received optical power inevitably decreases over distance due to attenuation and beam divergence.
Attenuation losses are minimal at meter-level ranges in air~\cite{denny1993}, while beam divergence is dependent on focus, which is fully configurable via the prototype's aspheric lens (Section \ref{subsec:track_steer_design}).

Our prototype's tracking range is dependent on the illumination strength of the LEDs used for retro-reflective imaging. As ambient light intensity or target distance increases, a greater strength is required to ensure sufficient light reaches the target and retroreflects back to cameras to distinguish the target from its background. %
We thus measure the LED power (both electrical and optical output) required for \name\ to isolate the retroreflective tag at varying distances, in both a bright room (744~lx) and darkened room (4~lx).

Figure \ref{fig:led-depth} shows that in the 4~lx case, the required LED power increases roughly linearly with the target's distance from the system. In the brighter lighting conditions, the relationship is exponential because LED light must both travel further and compete with ambient lighting. The maximum tracking distances are 3~m and 4~m for the 744~lx and 4~lx scenarios, respectively. This range compares favorably to existing work~\cite{iyer2018charging} which supports similarly sized, albeit stationary receivers at a maximum of 4.3~m. Notably, \name\ is compatible with any visual tracking approaches -- not just retroreflective imaging -- potentially increasing supported distance and eliminating illumination requirements.

\para{Power Consumption} Excluding the electrical power consumed by the laser ($\leq$12~W), the tracking and steering system consumes $\leq$37.4~W (25.4~W for the LED and stereo-cameras and $\leq$12~W for the mirror, whose consumption is dependent on steering speeds and thus robot motion). As a reference, an average laptop consumes around 65~W of power~\cite{laptop_power}.

\subsection{Low-Power Optical Communication Evaluation}
\label{subsec:comms_eval}

We evaluate the performance of \name's communication scheme under a variety of real-world deployment conditions, finding that it enables error-free communication under each.

\begin{figure*}[t]
\vspace{4pt} %
\captionsetup[subfigure]{oneside,margin={0.55cm,0cm}}
\begin{subfigure}[t]{.196\textwidth}
    \centering
    \includegraphics[width=\columnwidth]{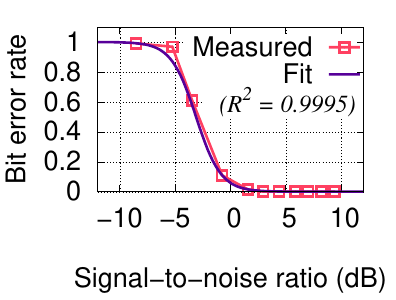}
    \caption{BER vs. SNR.}
    \label{fig:snr_ber}
\end{subfigure}
\captionsetup[subfigure]{oneside,margin={0.6cm,0cm}}
\begin{subfigure}[t]{.196\textwidth}
    \centering
    \includegraphics[width=\columnwidth]{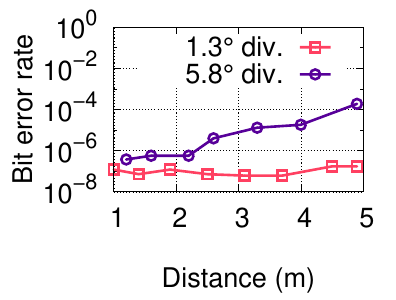}
    \caption{Distance.}
    \label{fig:ber_distance}
\end{subfigure}
\captionsetup[subfigure]{oneside,margin={0.6cm,0cm}}
\begin{subfigure}[t]{.196\textwidth}
    \centering
    \includegraphics[width=\columnwidth]{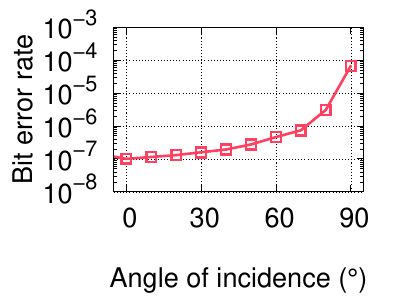}
    \caption{Incidence angle.}
    \label{fig:ber_angle}
\end{subfigure}
\captionsetup[subfigure]{oneside,margin={0.5cm,0cm}}
\begin{subfigure}[t]{.196\textwidth}
    \centering
    \includegraphics[width=\columnwidth]{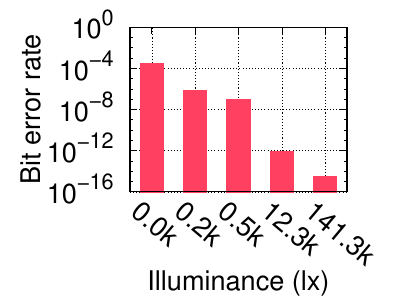}
    \caption{Ambient lighting.}
    \label{fig:ber_interference}
\end{subfigure}
\captionsetup[subfigure]{oneside,margin={0.5cm,0cm}}
\begin{subfigure}[t]{.196\textwidth}
    \centering
    \includegraphics[width=\columnwidth]{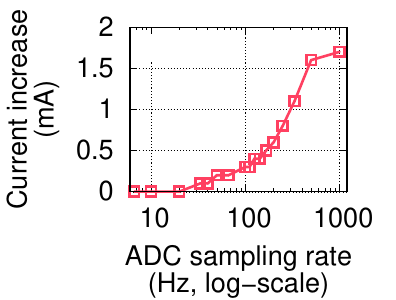}
    \caption{Power consumption.}
    \label{fig:comm_power}
\end{subfigure}
\caption{Low-power optical communication performance. (a) BER as a function of the signal-to-noise ratio of the received communication signal. The BER is consistently zero above the 3~dB point, meeting typical standards for wireless communication link quality~\cite{agrell2018information}. (b) BER vs. receiver distance for two beam divergences, specified in half-angle divergence. (c) BER as a function of incident angle onto the receiver photodiode. (d) Impact of ambient light interference, in terms of ambient light illuminance. (e) The power consumption of the decoding process (run on the robot) depends on the employed ADC sampling rate. Reported current increases are relative to 6.6 Hz consumption -- the minimum sampling rate required for MilliMobile locomotion.}
\end{figure*}

\para{Decoding Accuracy} Figure \ref{fig:snr_ber} plots the BER as a function of SNR obtained under twelve different optical powers and an analytical function fitted to the samples. The BER is consistently zero above the 3~dB point, corresponding to a signal strength of 16~mV. As a reference, in experiments with mobile robots in Section \ref{subsec:end_to_end}, we consistently measure signal strengths between 40~mV and 55~mV. 

\para{Target Depth} Figure \ref{fig:ber_distance} shows the BER as a function of receiver distance for two laser beam divergences. To obtain these results, we fix our laser along a horizontal track and move our communication PCB in increments of 0.5~m. At each distance, we record the SNR and compute the corresponding BER using our analytical fit in Figure \ref{fig:snr_ber}. In the low divergence case, we maintain a relatively constant irradiance across the entire tested space (with minor perturbations caused by beam nonuniformities) and thus observe a negligible increase in BER. In the larger divergence case, the decreasing optical density outweighs the effect of beam nonuniformities, and we observe a steadily increasing BER. For both divergences, we maintain a BER below pre-FEC (forward error correction) thresholds of $3.8\times10^{-3}$~\cite{agrell2018information}, indicating the signal can be decoded without errors.

\para{Laser Incident Angle} As a robot moves relative to the \name\ unit, the steered beam's angle of incidence onto its receiver will necessarily vary, potentially influencing BER as photodiodes possess varying angular responses. To examine this effect, we measure the received SNR across incident angles and convert these to BERs via our analytical function (Figure \ref{fig:snr_ber}). To isolate the impact of angle, we ensure the irradiance is 200 mW/cm$^2$ at each measurement. As shown in Figure \ref{fig:ber_angle}, though we observe a 3~dB power decrease and associated BER increase at approximately 80$\degree$,  BER remains under standard pre-FEC thresholds~\cite{agrell2018information} across angles.

\para{Ambient Light}  Ambient lighting conditions, induced by indoor luminaries or sunlight, can affect the communication link's noise floor, influencing SNR and thus BER. Figure \ref{fig:ber_interference} shows the BERs in five indoor and outdoor settings. Somewhat counterintuitively, environments with stronger ambient illumination exhibit lower BERs. This is because the noise floor is \emph{higher} in low-light levels as the photodiode (operating in photovoltaic mode) generates inconsistent current. In brighter environments, the photodiode generates stable photocurrent, reducing signal swings. In both cases, BER is sufficiently low to achieve zero post-correction errors. 

\para{Decoding Power Consumption} The power consumed by a robot's microprocessor to decode the laser signal is determined by its ADC sampling rate, which must be at least twice the highest transmitted frequency to satisfy the Nyquist–Shannon sampling theorem. Higher sampling rates expand the range of usable FSK frequencies for greater throughput, at the cost of greater energy consumption. 

Figure \ref{fig:comm_power} shows the additional current draw of decoding as a function of ADC sampling rate. The minimum sampling rate required for MilliMobile operation (absent communication) is 6.6~Hz. For the 100~Hz rate used in our communication evaluations, only 0.3~mA of additional current is drawn. For reference, MilliMobiles draw 4.5~mA current when moving at maximum speed with communications disabled. 
Thus, sampling at 100-Hz uses just 6.7\% of the current required for locomotion. Compared to Bluetooth Low Energy, which draws $\geq10$~mA to receive data at \emph{any} throughputs~\cite{johnson2023millimobile}, \name\ provides a lower-power alternative drawing 97\% less current for transmission of simple commands.

\subsection{End-to-End Demonstration}
\label{subsec:end_to_end}

\begin{figure*}[t]
\centering
\begin{subfigure}[t]{.31\textwidth}
    \centering
    \includegraphics[width=0.95\textwidth]{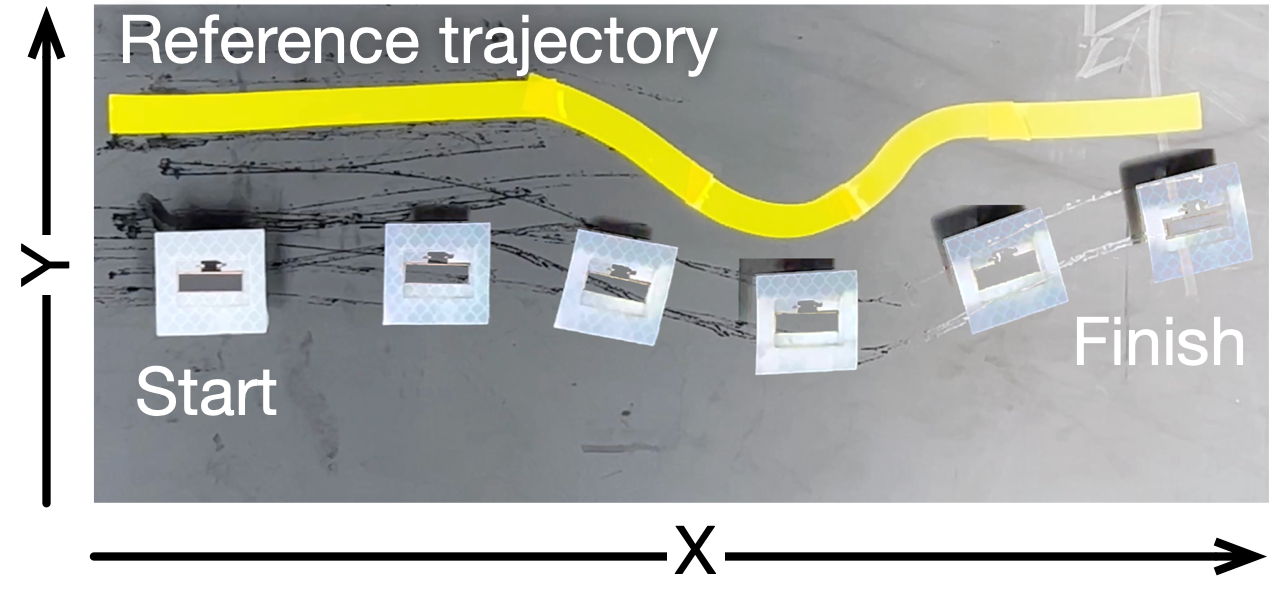}
    \\
    \includegraphics[width=1\textwidth]{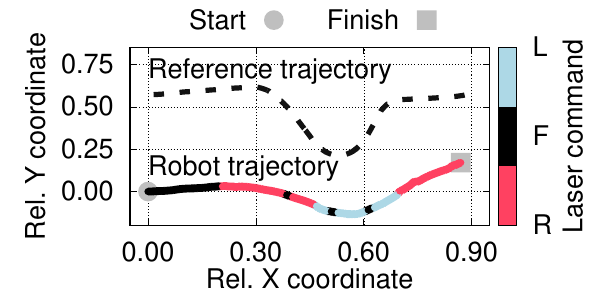}
    \caption{Path following.}
    \label{fig:path_following}
\end{subfigure}
\hfill
\begin{subfigure}[t]{.66\textwidth}
    \centering
    \hfill
    \begin{subfigure}[t]{0.47\columnwidth}
        \centering
        \includegraphics[width=.95\textwidth]{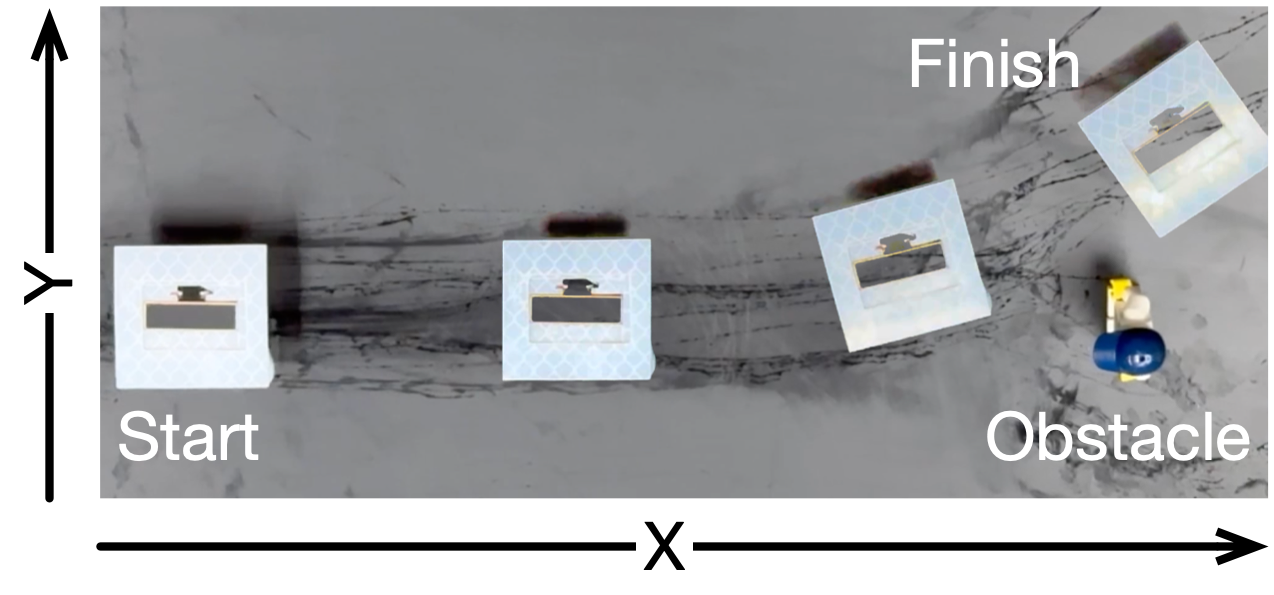}
        \\
        \includegraphics[width=1\textwidth]{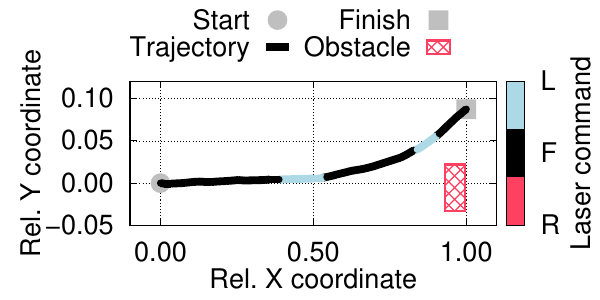}
    \end{subfigure}
    \hfill
    \begin{subfigure}[t]{0.47\columnwidth}
        \centering
        \includegraphics[width=.95\textwidth]{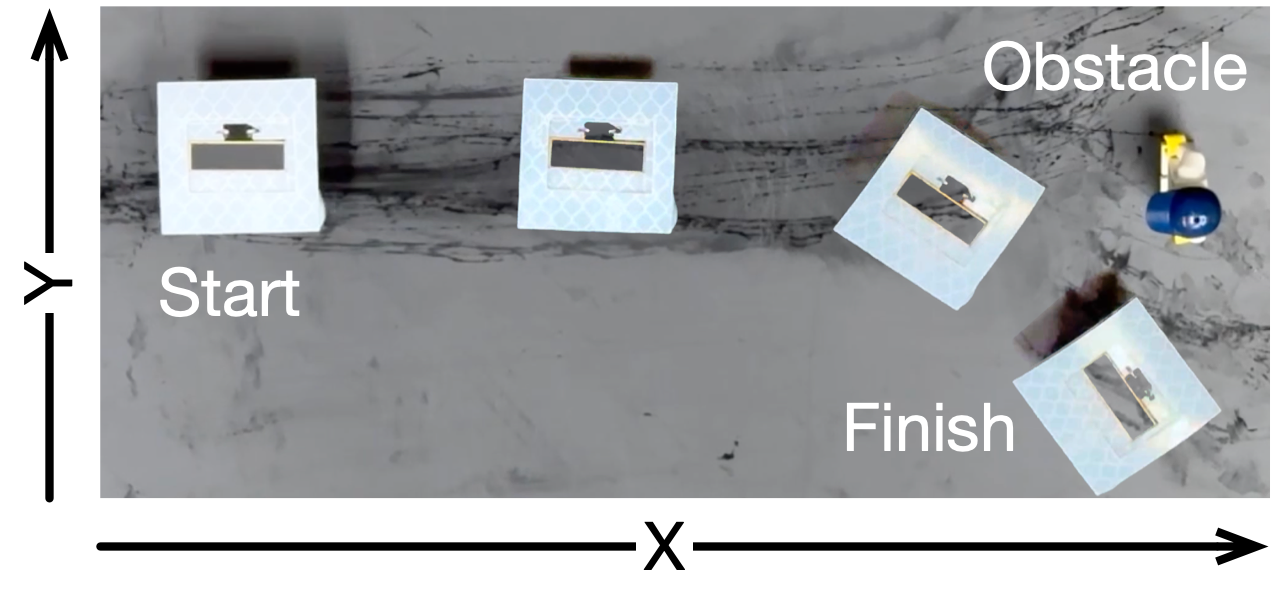}
        \\
        \includegraphics[width=1\textwidth]{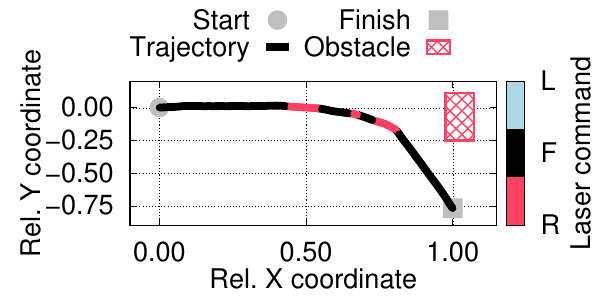}
    \end{subfigure}
    \caption{Obstacle avoidance.}
    \label{fig:obstacle_avoidance}
\end{subfigure}
\caption{End-to-end evaluation using \name\ to power and control MilliMobile robots. (a) \name\ moves the MilliMobile along a path, marked by the yellow tape. Bottom row shows the ground truth trajectory relative to the reference path. The robot maintains a consistent offset even as it moves around the curve. (b) Laser-powered MilliMobile navigating around an obstacle in the scene via laser-provided commands. Timelapse images (top) show the robot's trajectory in two steering cases (both left and right). The laser command sent to the robot is either \texttt{L} for left, \texttt{R} for right, or \texttt{F} for forward. }
\vspace{-10pt}
\end{figure*}

We deploy \name\ to deliver power and commands via the optical channel to MilliMobile robots. Prior MilliMobile deployments utilized diffuse light or omnidirectional radio frequency waves for power, with Bluetooth Low Energy for communication. We show \name\ nearly doubles the locomotion speed of MilliMobiles from their speed under sunlight, providing optical power density of up to 200 mW/cm$^2$. Additionally, laser-provided commands reliably control the trajectory of robots and guide them around obstacles. 

To assess the robots' trajectories and adherence to steering commands, we capture overhead video of the moving robot and synchronize it with commands sent at each timestamp. We also obtain the robot's pixel coordinates (x, y) at each timestamp, normalized by its start and maximum (x, y) pixel.

We first demonstrate \name's ability to consistently move the robot along a predetermined path. As shown in Figure \ref{fig:path_following}, we place yellow electrical tape along the testbed surface
with a slight curve in the center. The robot starts with an offset relative to the tape to keep it visible throughout the robot's motion. We then power on \name\ and manually send wireless laser commands instructing the robot to move forward, left, or right to maneuver it along the trajectory. We plot the relative (x, y) coordinate of the robot as it moves along the reference trajectory, in addition to the corresponding steering commands. The robot starts with a relative y-offset of 0.57. Throughout its motion, it maintains a median pixel offset of 0.51 and mean offset of 0.49, a marginal 11\% and 15\% difference, respectively. This variation can be explained by mechanical irregularities that caused the robot's wheels to make inconsistent contact with the ground.

Additionally, we deploy \name\ to steer robots around an obstacle in the scene. Figure \ref{fig:obstacle_avoidance} shows an obstacle aligned with the robot along the x-axis. When we operate \name\ sending only forward (\texttt{F}) commands, the robot collides with the obstacle, as expected. When provided commands to steer to either the left or right of the obstacle, it avoids collision.

Across both of these deployment scenarios, only 0.3~mA of current produced from harvesting laser light is consumed for decoding optical communication signals. While receiving data, the MilliMobile moves with an average speed of 1~cm/s, an 82\% increase compared to the top speed of 5.5~mm/s reported in prior work under close-range high-power LED illumination with no communication capabilities~\cite{johnson2023millimobile} .

\section{Conclusion and Future Work}
\name\  provides concurrent power and communication to mobile robots using narrow-beam laser light. We show that \name\ can maintain beam alignment and establish error-free communication to robotic targets moving arbitrarily in 3D space, at up to 4~m distances. Hardware demonstrations with battery-free MilliMobile robots show that our laser-based system achieves nearly 2x improvements in average speeds during path following and obstacle avoidance tasks. Beyond supporting battery-free robots, \name\ can be used for continuous re-charging to augment robot run-times.

In terms of future work, while our experiments focused on a single-robot environment, \name\ could enable swarms of robots for various advanced applications. %
\name's functionality can also be extended with higher-throughput optical communication schemes to support richer command sets and additional robot tracking algorithms to accommodate higher robot speeds. %
Finally, we hope to explore technical solutions to ensuring laser safety in real-world \name\ deployments. %

\bibliographystyle{bibs/IEEEtran}
\bibliography{bibs/aliases,bibs/references}

\begin{thebibliography}{10}
\providecommand{\url}[1]{#1}
\csname url@rmstyle\endcsname
\providecommand{\newblock}{\relax}
\providecommand{\bibinfo}[2]{#2}
\providecommand\BIBentrySTDinterwordspacing{\spaceskip=0pt\relax}
\providecommand\BIBentryALTinterwordstretchfactor{4}
\providecommand\BIBentryALTinterwordspacing{\spaceskip=\fontdimen2\font plus
\BIBentryALTinterwordstretchfactor\fontdimen3\font minus
  \fontdimen4\font\relax}
\providecommand\BIBforeignlanguage[2]{{%
\expandafter\ifx\csname l@#1\endcsname\relax
\typeout{** WARNING: IEEEtran.bst: No hyphenation pattern has been}%
\typeout{** loaded for the language `#1'. Using the pattern for}%
\typeout{** the default language instead.}%
\else
\language=\csname l@#1\endcsname
\fi
#2}}

\bibitem{kabir2023internet}
H.~Kabir, M.-L. Tham, and Y.~C. Chang, ``Internet of robotic things for mobile
  robots: concepts, technologies, challenges, applications, and future
  directions,'' \emph{Digital Communications and Networks}, vol.~9, no.~6, pp.
  1265--1290, 2023.

\bibitem{thackeray2012electrical}
M.~M. Thackeray, C.~Wolverton, and E.~D. Isaacs, ``Electrical energy storage
  for transportation—approaching the limits of, and going beyond, lithium-ion
  batteries,'' \emph{Energy \& Environmental Science}, vol.~5, no.~7, pp.
  7854--7863, 2012.

\bibitem{ma2015design}
K.~Y. Ma, P.~Chirarattananon, and R.~J. Wood, ``Design and fabrication of an
  insect-scale flying robot for control autonomy,'' in \emph{Proc. of {IROS}},
  2015, pp. 1558--1564.

\bibitem{james2018liftoff}
J.~James, V.~Iyer, Y.~Chukewad, S.~Gollakota, and S.~B. Fuller, ``Liftoff of a
  190 mg laser-powered aerial vehicle: The lightest wireless robot to fly,'' in
  \emph{Proc. of {ICRA}}, 2018, pp. 3587--3594.

\bibitem{fuller2019four}
S.~B. Fuller, ``Four wings: An insect-sized aerial robot with steering ability
  and payload capacity for autonomy,'' \emph{IEEE Robotics and Automation
  Letters}, vol.~4, no.~2, pp. 570--577, 2019.

\bibitem{bhushan2018milligram}
P.~Bhushan and C.~J. Tomlin, ``Milligram-scale micro aerial vehicle design for
  low-voltage operation,'' in \emph{Proc. of {IROS}}, 2018, pp. 1--9.

\bibitem{jafferis2019untethered}
N.~T. Jafferis, E.~F. Helbling, M.~Karpelson, and R.~J. Wood, ``Untethered
  flight of an insect-sized flapping-wing microscale aerial vehicle,''
  \emph{Nature}, vol. 570, no. 7762, pp. 491--495, 2019.

\bibitem{johnson2023millimobile}
K.~Johnson, Z.~Englhardt, V.~Arroyos, D.~Yin, S.~Patel, and V.~Iyer,
  ``Millimobile: An autonomous battery-free wireless microrobot,'' in
  \emph{Proceedings of the 29th Annual International Conference on Mobile
  Computing and Networking}, 2023, pp. 1--16.

\bibitem{ozaki2021wireless}
T.~Ozaki, N.~Ohta, T.~Jimbo, and K.~Hamaguchi, ``A wireless
  radiofrequency-powered insect-scale flapping-wing aerial vehicle,''
  \emph{Nature Electronics}, vol.~4, no.~11, pp. 845--852, 2021.

\bibitem{elkunchwar2021toward}
N.~Elkunchwar, S.~Chandrasekaran, V.~Iyer, and S.~B. Fuller, ``Toward
  battery-free flight: Duty cycled recharging of small drones,'' in \emph{Proc.
  of {IROS}}.\hskip 1em plus 0.5em minus 0.4em\relax IEEE, 2021, pp.
  5234--5241.

\bibitem{shen2024sunlight}
W.~Shen, J.~Peng, R.~Ma, J.~Wu, J.~Li, Z.~Liu, J.~Leng, X.~Yan, and M.~Qi,
  ``Sunlight-powered sustained flight of an ultralight micro aerial vehicle,''
  \emph{Nature}, vol. 631, no. 8021, pp. 537--543, 2024.

\bibitem{liller2025development}
J.~Liller, R.~Goel, A.~Aziz, J.~Hester, and P.~Nguyen, ``Development of a
  battery free, solar powered, and energy aware fixed wing unmanned aerial
  vehicle,'' \emph{Scientific Reports}, vol.~15, no.~1, p. 6141, 2025.

\bibitem{sun_irradiance}
ISO, ``{IEC-60904-3 Photovoltaic devices - Part 3: Measurement principles for
  terrestrial photovoltaic (PV) solar devices with reference spectral
  irradiance data},'' \url{https://www.iso.org/standard/17723.html}, 2019.

\bibitem{carver2024catch}
C.~J. Carver, H.~Schwartz, Q.~Shao, N.~Shade, J.~Lazzaro, X.~Wang, J.~Liu,
  E.~Fossum, and X.~Zhou, ``Catch me if you can: Laser tethering with highly
  mobile targets,'' in \emph{Proc. of {NSDI}}, 2024, pp. 1847--1865.

\bibitem{chen2019controlled}
Y.~Chen, H.~Zhao, J.~Mao, P.~Chirarattananon, E.~F. Helbling, N.-s.~P. Hyun,
  D.~R. Clarke, and R.~J. Wood, ``Controlled flight of a microrobot powered by
  soft artificial muscles,'' \emph{Nature}, vol. 575, no. 7782, pp. 324--329,
  2019.

\bibitem{wu2019insect}
Y.~Wu, J.~K. Yim, J.~Liang, Z.~Shao, M.~Qi, J.~Zhong, Z.~Luo, X.~Yan, M.~Zhang,
  X.~Wang, \emph{et~al.}, ``Insect-scale fast moving and ultrarobust soft
  robot,'' \emph{Science Robotics}, vol.~4, no.~32, p. eaax1594, 2019.

\bibitem{hutama2021millimeter}
R.~Y. Hutama, M.~M. Khalil, and T.~Mashimo, ``A millimeter-scale rolling
  microrobot driven by a micro-geared ultrasonic motor,'' \emph{IEEE Robotics
  and Automation Letters}, vol.~6, no.~4, pp. 8158--8164, 2021.

\bibitem{blackwell2005recent}
T.~Blackwell, ``Recent demonstrations of laser power beaming at {DFRC} and
  {MSFC},'' in \emph{AIP Conference Proceedings}, vol. 766.\hskip 1em plus
  0.5em minus 0.4em\relax American Institute of Physics, 2005, pp. 73--85.

\bibitem{jin2018wireless}
K.~Jin and W.~Zhou, ``Wireless laser power transmission: A review of recent
  progress,'' \emph{IEEE Trans. on Power Electronics}, vol.~34, no.~4, pp.
  3842--3859, 2018.

\bibitem{iyer2018charging}
V.~Iyer, E.~Bayati, R.~Nandakumar, A.~Majumdar, and S.~Gollakota, ``Charging a
  smartphone across a room using lasers,'' \emph{Proc. of {IMWUT}.}, vol.~1,
  no.~4, pp. 1--21, 2018.

\bibitem{sheng2022adaptive}
Q.~Sheng, J.~Geng, Z.~Chang, A.~Wang, M.~Wang, S.~Fu, W.~Shi, and J.~Yao,
  ``Adaptive wireless power transfer via resonant laser beam over large dynamic
  range,'' \emph{IEEE Internet of Things Journal}, vol.~10, no.~10, pp.
  8865--8877, 2022.

\bibitem{12hflight2011lasermotive}
M.~C. Achtelik, J.~Stumpf, D.~Gurdan, and K.-M. Doth, ``{Design of a flexible
  high performance quadcopter platform breaking the MAV endurance record with
  laser power beaming},'' in \emph{Proc. of {IROS}}, 2011, pp. 5166--5172.

\bibitem{blackwell2003nasa}
T.~Blackwell, ``{Recent demonstrations of Laser power beaming at DFRC and
  MSFC},'' \emph{AIP Conference Proceedings}, vol. 766, no.~1, pp. 73--85,
  2005.

\bibitem{milanovic2011memseye}
V.~Milanovi{\'c}, N.~Siu, A.~Kasturi, M.~Radoji{\v{c}}i{\'c}, and Y.~Su,
  ``{MEMSEye for optical 3D position and orientation measurement},'' in
  \emph{MOEMS and Miniaturized Systems}, vol. 7930, 2011, pp. 254--259.

\bibitem{sofka2009laser}
J.~Sofka, V.~V. Nikulin, V.~A. Skormin, D.~H. Hughes, and D.~J. Legare, ``Laser
  communication between mobile platforms,'' \emph{{Transactions on Aerospace
  and Electronic Systems}}, vol.~45, no.~1, pp. 336--346, 2009.

\bibitem{llcdpat}
J.~Burnside, S.~Conrad, A.~Pillsbury, and C.~DeVoe, ``Design of an inertially
  stabilized telescope for the {LLCD},'' in \emph{Free-Space Laser
  Communication Technologies}, vol. 7923, 2011, pp. 133--140.

\bibitem{abadiautolocatingfso}
M.~M. Abadi, M.~A. Cox, R.~E. Alsaigh, S.~Viola, A.~Forbes, and M.~P. Lavery,
  ``A space division multiplexed free-space-optical communication system that
  can auto-locate and fully self align with a remote transceiver,''
  \emph{Scientific Reports}, vol.~9, no.~1, pp. 1--8, 2019.

\bibitem{ishola2021pat}
F.~Ishola and M.~Cho, ``{Experimental Study on Photodiode Array Sensor Aided
  MEMS Fine Steering Mirror Control for Laser Communication Platforms},''
  \emph{IEEE Access}, vol.~9, pp. 100\,197--100\,207, 2021.

\bibitem{gupta2022cyclops}
H.~Gupta, M.~Curran, J.~Longtin, T.~Rockwell, K.~Zheng, and M.~Dasari,
  ``{Cyclops: An FSO-based wireless link for VR headsets},'' in \emph{Proc. of
  {SIGCOMM}}, 2022, pp. 601--614.

\bibitem{kasturi2016uav}
A.~Kasturi, V.~Milanovic, B.~H. Atwood, and J.~Yang, ``{UAV-borne lidar with
  MEMS mirror-based scanning capability},'' in \emph{Laser Radar Technology and
  Applications XXI}, vol. 9832, 2016, pp. 206--215.

\bibitem{carver2024demonstration}
C.~J. Carver, T.~Itagaki, K.~Liu, M.~G. Manik, Z.~Englhardt, V.~Iyer, and
  X.~Zhou, ``Demonstration of laser power delivery for mobile microrobots,'' in
  \emph{Proceedings of the 10th Workshop on Micro Aerial Vehicle Networks,
  Systems, and Applications}, 2024, pp. 19--24.

\bibitem{roberts2016overview}
W.~Roberts, D.~Antsos, A.~Croonquist, S.~Piazzolla, L.~Roberts~Jr,
  V.~Garkanian, T.~Trinh, M.~Wright, R.~Rogalin, J.~Wu, \emph{et~al.},
  ``{Overview of ground station 1 of the NASA space communications and
  navigation program},'' in \emph{Free-Space Laser Communication and
  Atmospheric Propagation}, vol. 9739, 2016, pp. 82--99.

\bibitem{abdelfatah2022review}
R.~Abdelfatah, N.~Alshaer, and T.~Ismail, ``{A review on pointing, acquisition,
  and tracking approaches in UAV-based FSO communication systems},''
  \emph{Optical and Quantum Electronics}, vol.~54, no.~9, 2022.

\bibitem{walsh2022demonstration}
S.~M. Walsh, S.~F. Karpathakis, A.~S. McCann, B.~P. Dix-Matthews, A.~M. Frost,
  D.~R. Gozzard, C.~T. Gravestock, and S.~W. Schediwy, ``{Demonstration of 100
  Gbps coherent free-space optical communications at LEO tracking rates},''
  \emph{Scientific Reports}, vol.~12, no.~1, 2022.

\bibitem{tsai1987versatile}
R.~Tsai, ``{A versatile camera calibration technique for high-accuracy 3D
  machine vision metrology using off-the-shelf TV cameras and lenses},''
  \emph{IEEE J. Robotics Autom.}, vol.~3, no.~4, pp. 323--344, 1987.

\bibitem{OpenCV}
OpenCV, \url{https://opencv.org/}, 2024.

\bibitem{hartley1997triangulation}
R.~I. Hartley and P.~Sturm, ``Triangulation,'' \emph{Computer Vision and Image
  Understanding}, vol.~68, no.~2, pp. 146--157, 1997.

\bibitem{distortion_optotune_mre3_manual}
Optotune, ``{MR-E-3 Development Kit},'' 2025.

\bibitem{hoff1989surfaces}
W.~Hoff and N.~Ahuja, ``Surfaces from stereo: Integrating feature matching,
  disparity estimation, and contour detection,'' \emph{IEEE Trans. Pattern
  Anal. Mach. Intell.}, vol.~11, no.~2, pp. 121--136, 1989.

\bibitem{kim2024design}
H.~Kim, Y.~Jeon, and M.~G. Lee, ``Design and control of fast steering mirror
  for satellite communication process,'' in \emph{Intl. Conference on Control,
  Automation and Information Sciences}, 2024, pp. 1--6.

\bibitem{jeromin1986m}
L.~Jeromin and V.~Chan, ``{M-ary FSK performance for coherent optical
  communication systems using semiconductor lasers},'' \emph{IEEE Transactions
  on Communications}, vol.~34, no.~4, pp. 375--381, 1986.

\bibitem{denny1993}
M.~Denny, \emph{Air and Water}.\hskip 1em plus 0.5em minus 0.4em\relax
  Princeton University Press, 1993.

\bibitem{laptop_power}
Jackery, ``How many watts does a laptop use: Macbook, dell, asus and more,''
  \url{https://www.jackery.com/blogs/knowledge/how-many-watts-a-laptop-uses},
  2023.

\bibitem{agrell2018information}
E.~Agrell and M.~Secondini, ``Information-theoretic tools for optical
  communications engineers,'' in \emph{IEEE Photonics Conference}, 2018.

\end{thebibliography}

\end{document}